\documentclass[runningheads]{llncs}
\usepackage[T1]{fontenc}
\usepackage{graphicx}
\usepackage{amsmath}
\usepackage{amssymb}
\usepackage{multirow}
\usepackage{bbding}
\begin{document}
%
% \title{Improving accuracy of medical VQA problems based on QueryVQA}
\title{Q2ATransformer: Improving Medical VQA via an Answer Querying Decoder}

\author{Yunyi Liu\inst{1}\and
Zhanyu Wang\inst{1}\and
Dong Xu\inst{2} \and
Luping Zhou\inst{1}}
\authorrunning{Y. Liu. et al.}

\institute{The University of Sydney, Sydney,NSW,Australia\\
\email{\{yunyi.liu,zhanyu.wang,luping.zhou\}@sydney.edu.au}\\
 \and
The University of Hong Kong, Hong Kong SAR\\
\email{dongxu@cs.hku.hk}}

\maketitle              % typeset the header of the contribution
\begin{abstract}
Medical Visual Question Answering (VQA) systems play a supporting role to understand clinic-relevant information carried by medical images. The questions to a medical image include two categories: close-end (such as Yes/No question) and open-end.  To obtain answers, the majority of the existing medical VQA methods rely on classification approaches, while a few works attempt to use generation approaches or a mixture of the two to process the two kinds of questions separately (classification for the close-end and generation for the open-end). The classification approaches are relatively simple but perform poorly on long open-end questions, while the generation approaches face the challenge of generating many non-existent answers, resulting in low accuracy rates.   To bridge this gap, in this paper, we propose a new Transformer based framework for medical VQA (named as Q2ATransformer), which integrates the advantages of both the classification and the generation approaches and provides a unified treatment for the close-end and open-end questions. Specifically, we introduce an additional Transformer decoder with a set of learnable candidate answer embeddings to query the existence of each answer class to a given image-question pair. Through the Transformer attention, the candidate answer embeddings interact with the fused features of the image-question pair to make the decision.  In this way, despite being a classification-based approach, our method provides a mechanism to interact with the answer information for prediction like the generation-based approaches. On the other hand, by classification, we mitigate the task difficulty by reducing the search space of answers. Our method achieves new state-of-the-art performance on two medical VQA benchmarks.  Especially, for the open-end questions, we achieve 79.19\% on VQA-RAD and 54.85\% on PathVQA, with 16.09\% and 41.45\% absolute improvements, respectively.

\keywords{Medical VQA  \and Attention Mechanism \and Classification.}
\end{abstract}
%
%
%-----------------------------------------------------------~--------------------------------------------------------------------------
\section{Introduction}
Visual question answering (VQA) is known to be a challenging AI task that answers image-related questions based on image content. This process involves both image and natural language processing techniques and usually comprises of four key components: extracting image features, extracting question features, integrating features, and answering. Recent years have witnessed significant progress in this field \cite{jiang2020defense,wu2019differential}. Medical VQA is a natural extension of VQA to medical images accompanied by clinic-relevant questions. Through questioning and answering, it offers a user-friendly way to assist clinic decisions. The questions in medical VQA could be either close-end, such as Yes/No questions, or open-end.

Medical VQA is still in its early stage of development and the current performance is far from being satisfying. Most existing methods \cite{nguyen2019overcoming,do2021multiple,he2020pathvqa,finn2017model,eslami2021does} could be referred to as closed-type approaches, as illustrated in Figure 1 (a), which treat each answer as a class and apply a classification model directly to the fused features of the input image-question pair to predict answers. The advantage of such approaches is that by treating VQA as classification tasks, they reduce the complexity of the task and make the answer search space smaller. Despite the good performance on Yes/No questions, closed-type approaches are difficult to accurately predict the answer for open-end questions that are much longer and more varied than the close-end ones. On the other hand, a few works \cite{ambati2018sequence,khare2021mmbert} treat VQA as a generation task and employ generation-based approaches to produce answers word by word. They are referred to as the open-type approaches in Figure 1(b). In these approaches, current word generation usually depends on previous words of the answer. Therefore, these approaches allow the image-question features to interact with the answer information for the prediction, potentially improving the long answer prediction. However, due to the tremendous search space of the generated answers, these approaches tend to produce many non-existent answers, leading to low accuracy rates, therefore are not currently the mainstream of medical VQA. Although there are attempts to combine these two types of approaches \cite{ren2020cgmvqa}, they straightforwardly treat close-end and open-end questions separately, e.g., classification for close-end ones and generation for open-end ones. 

To bridge the research gap and promote medical VQA, we introduce a new model framework Q2ATransformer and refer it to as semi-open type, as shown in Figure 1 (c). By semi-open, we keep adopting classification-based approaches to make the answer search space small, and at the same time introduce the learning of answer semantics so that the fused image-question features and the answer semantics could interact for better prediction, like the generation-based approaches. Our model mitigates the shortcomings of the classification-based closed-type framework while enjoying the advantages of the generation-based open-type framework. To achieve this, we introduce a set of learnable candidate answer embeddings and let the image-question feature interact with the candidate answer embeddings by sending them through a transformer decoder. In the decoder, the candidate answer embeddings work as a query to calculate their relationships with the fused image-question features to decide the existence of the answer classes. By this, our classification considers the interaction of answer information and the fused image-question features, which is different from the existing classification-based approaches. Compared with the generation-based open-type approaches, our model reduces the task difficulty and significantly improves the accuracy rates.

Last but not the least, our model provides a uniform treatment for both the close-end and the open-end questions.

The main contribution of this paper could be summarized as follows.

First, we proposed a framework of semi-open type for medical VQA, which bridges the advantages of both the classification-based closed-type framework and the generation-based open-type frameworks in medical VQA literature. This is achieved by a designed mechanism to learn and make use of candidate answer embedding through a transformer decoder while limiting the search scope of answers through classification.

Second, we proposed a Cross-modality Fusion Network (CMAN) to effectively fuse the image and question features. It directly concatenates the two modal features instead of conducting matrix multiplication or summation for feature fusion to mitigate information loss. Then the relations between the image and question features are captured through computing self-attention on the concatenated features to produce the fused features. CMAN outperforms the commonly used image-question fusion methods in medical VQA as shown in our ablation study.

Third, our model demonstrates superior performance on two large medical VQA benchmarks for both close-end and open-end questions. Especially, our improvement on open-end question answering is overwhelming, with 16\% and 41\% absolute improvements on VQA-RAD and PathVQA, respectively, verifying the effectiveness of our proposed semi-open framework.

\begin{figure}[t]
\includegraphics[width=\textwidth]{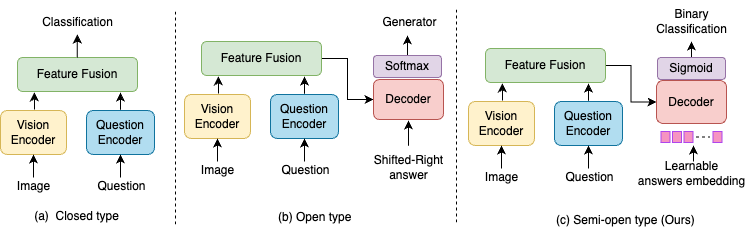}
\caption{Paradigms for medical VQA frameworks. (a) Closed-type framework treats VQA as predicting answer classes, where a classifier is built directly on top of the fused image-question features. (b) Open-type framework is generation-based, where the fused image-question features interact with the previous words of the answer to generate the next word of the answer through a text decoder. (c) Our proposed semi-open framework learns candidate answer embeddings through a decoder, where they interact with the fused image-question features to improve the prediction of answer classes. 
} \label{firstPage}
\end{figure}

\section{Method}~\label{method}
In this section, we present Q2ATransformer, a semi-open structured model for medical VQA. We first give an overview of our model, and then describe our Visual-Question Encoder in Sec. 3.1 and Answer Querying Decoder in Sec. 3.2. 

\begin{figure}[h]
\includegraphics[width=\textwidth]{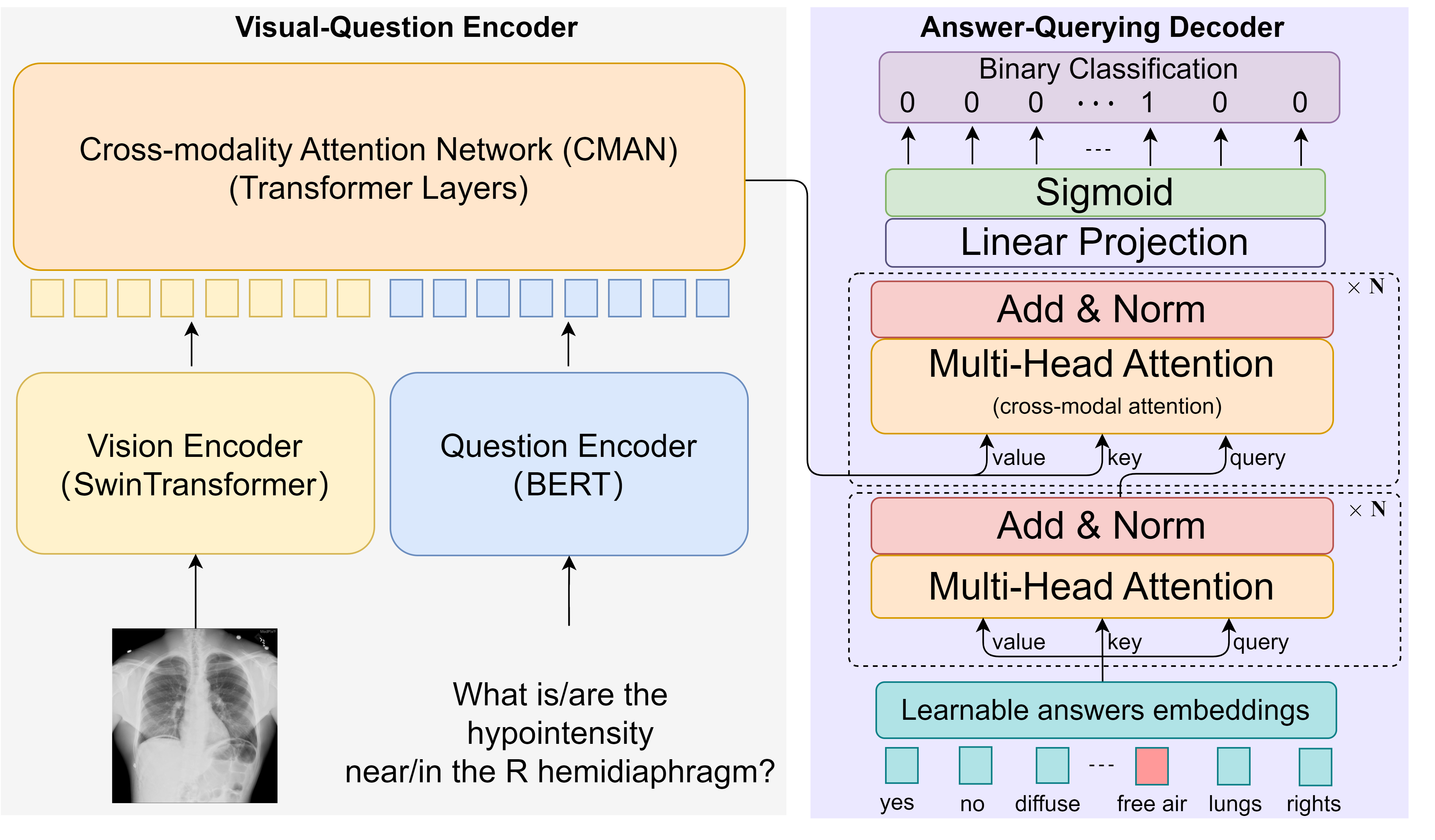}
\caption{Overview of Q2ATransformer. The input image-question pair is sent to a Visual-Question Encoder to extract and fuse image and question features. The Visual-Question Encoder consists of a Swin-Transformer-based vision encoder, a BERT-based question encoder, and a proposed Cross-modality Attention Network for feature fusion. The fused feature proceeds to the Answer-Querying Decoder, where the input learnable candidate answer embeddings are utilized as the query to compute the attention map and refined according to the attended fused image-question features to predict the presence of the queried answers.
%After combining the spatial features of the extracted input image question pairs, each answer embedding is sent to the Transformer decoder for querying (by comparing the answer embedding with the features at each spatial location to generate an attention map) and adaptively pooling the desired features (by linearly combining spatial features based on the attention map). The pooled features are then used to predict the presence of the queried answers.
} \label{fig:framework}
\end{figure}

 An overview of our proposed Q2ATransformer model is given in Fig.~\ref{fig:framework}. It follows the majority of medical VQA methods to predict answer classes but exploits candidate answer embeddings for the prediction. Q2ATransformer consists of a Visual-Question Encoder and an Answer-Querying Decoder. The Visual-Question Encoder takes a medical image and a clinic-relevant question as the input and outputs a fused feature with both image and question information. It consists of three parts: vision encoder, question encoder, and fusion network. We use Swin transformer as our vision encoder and BERT as the question encoder. For the fusion network, we propose a Cross-modality Attention Network (CMAN) to integrate image and question features. The Answer Querying Decoder takes the fused image-question feature and learnable candidate answer embeddings as the input and outputs the probability of each candidate answer. Our Answer Querying Decoder consists of two layers of transformer decoders and a classifier to make predictions.

\vspace{-5mm}
\subsection{Visual-Question Encoder} 
The Visual-Question Encoder consists of an image encoder, a question encoder, and a feature fusion module, elaborated as follows.

% The expression of the Visual-Question Encoder is as equation 1:\\
% \begin{equation}
%     F_{fusion} = \mathrm {CTAN} (Concat\left ( F_{q} ; F_{i} \right )) 
% \end{equation}

% Where CTAN is our cross transformer attention decoder, and concat is use to concatenate visual feature and text feature togethor. 

\textbf{Image Encoder.} 
Our encoder uses the Swin Transformer~\cite{liu2021swin} rather than CNN-based model as our image feature extractor. The advantages of Swin Transformer are three-fold. \underline{First}, Swin Transformer makes a vision transformer to a hierarchical structure as CNN, which can make the vision transformer more flexible at various scales and has linear computational complexity with the increase of image size. \underline{Second}, Swin Transformer considers cross-window connection through window shift to obtain long-range dependencies, which introduces more interactions between grids. Therefore, it can provide more regional features and interactions compared with CNN, which is more suitable for the fine-grained nature of medical images. \underline{Third}, Swin Transformer was pretrained on a large dataset, so it is a very robust feature extractor. 
Based on these characteristics, we choose Swin Transformer to encode our input image.

Given an input image $\mathbf{I} \in \mathbb{R}^{H \times W \times C}$, where C is the number of channels and H and W stand for image height and width, respectively, the image embeddings $\mathbf{F}_i \in \mathbb{R}^{N \times D_f}$ can be expressed as $\mathbf{F}_i = \mathbf{W}_i \times \mathrm{SwinTransformer(\mathbf{I})} + \mathbf{b}_i$, where $\mathbf{W}_i$ and $\mathbf{b}_i$ are learnable parameters to project the output of Swin Transformer into the same dimension $D_f$ as the question embeddings. They also provide certain flexibility to adapt Swin Transformer to the datasets in our task. Here $N$ is the number of the extracted image regional features.
% We stacked another fully connected layer to match image feature with same dimension of text embedding, $\mathbf{F}_i = \mathrm{LN}\left ( \mathrm{SwinTransformer}\left ( \mathbf{I} \right )  \right ) $
% $\mathbf{F}_i \in \mathbb{R}^{(P \times P) \times D} $
% Then, we feed the image to Swin-transformer~\cite{liu2021swin} to extract the feature $\mathbf{F}_i \in \mathbb{R}^{(P \times P) \times D} $, where P represent the size of the feature map, and D denotes the dimension of the features.  \\
% where $H \times W$ represent the height and weight of the input image, then extract its spatial features $\mathbf{F}_i \in \mathbb{R}^{H0 \times W0 \times d0}$ using the Swin-transformer, where $\mathbf{H0\times W0}$ represent the height and weight of the feature map respectively, and $\mathbf{d}$ denotes the dimension of features.  \\
% \begin{equation}
%     \mathbf F_{i} = \mathrm{SwinTransformer}\left ( I \right ) 
% \end{equation}

\textbf{Question Encoder.} 
%BERT 的描述，结构是什么，BERT 的intro里面找几句总结的话
For the input question, we use the pre-trained BERT model~\cite{devlin2018bert} as the encoder to extract text features.
%As described in \cite{devlin2018bert}, BERT stands for Bidirectional Encoder Representations from Transformers, which is a pre-training model for language understanding. 
BERT~\cite{devlin2018bert} is a successful NLP model. It incorporates context from both directions of a sentence when embedding questions. It has been applied to question answering tasks with the state-of-the-art results, and is therefore chosen in our task as the question encoder. The question embeddings $\mathbf{F}_q \in \mathbb{R}^{M \times D_f}$ is obtained by $ \mathbf{F}_{q} = \mathrm {BERT}\left ( \mathbf{Q}_e \right ) $, where ${\mathbf Q}_e$ denotes the input question and $M$ is the question feature number and $D_f$ the question feature dimension. 
% Given an questions $\mathbf{Q}$ as input and tokenize them as ${\mathbf{Q}(t_1), \cdots, \mathbf{Q}(t_n)}$ , then input the question tokens into BERT Model to get the output of question embedding $\mathbf{F}q: [CLS], \mathbf{F}q(1), …, \mathbf{F}q(n)$, $\mathbf{F}_q \in \mathbb{R}^{H1}$, $H_1 represent word embedding diminution.

% \begin{equation}
% \mathbf{F}_{q} = \mathrm {BERT}\left ( \mathbf{Q} \right ) 
% \end{equation}

\textbf{Feature Fusion Mechanism.} 
After the image and question features are extracted, respectively, we propose the Cross-modality Attention Network (CMAN) to fuse the information from these two modalities. As medical images are fine-grained and the visual differences of clinical importance are often subtle, we explore a sophisticated way for feature fusion by investigating the interactions between image regional features and question features. In our proposed fusion module CMAN, 
%We wish to get the fused feature that has both information from medical images and relevant questions, and also we required the fused features have interactive information between image and question features. 
we first integrate the image features $\mathbf{F}_i$ and the question features $\mathbf{F}_q$ by concatenating them together. Compared with the commonly used matrix multiplication or summation for feature fusion, concatenation could mitigate information loss and facilitate the subsequent computation of image-question interaction in our module. After that, the concatenated features are passed to two transformer encoder layers to calculate the relationship between every pair of image question features through the self-attention mechanism of the Transformer. In this way, we could obtain the fused feature carrying the relation of image question features with minimal information loss. Mathematically, the fused feature ${\mathbf F}_f$ is obtained as follows.
\begin{equation}\label{Eqn:fusion}
    \begin{aligned}
        &{\mathbf F}_{c} = [{\mathbf F}_i; {\mathbf F}_q]\\
        &{\mathbf Q}_{F_c}={\mathbf W}_{q} {\mathbf F}_{c},~~ {\mathbf K}_{F_c}={\mathbf W}_{k} {\mathbf F}_{c},~~{\mathbf V}_{F_c}={\mathbf W}_{v} {\mathbf F}_{c}  \\
    &{\mathbf F}_{att}=\mathbf {Att({\mathbf Q}_{F_c}, {\mathbf K}_{F_c},{\mathbf V}_{F_c})} = \mathbf { softmax(\frac{{\mathbf Q}_{F_c} {\mathbf K}_{F_c}^T}{\sqrt{d_k}}){\mathbf V}_{F_c}} \\
    &{\mathbf F}_f = {\mathbf W}_f {\mathbf F}_{att} + {\mathbf b}_f
    \end{aligned}
\end{equation}
Here ${\mathbf W}_{q}$, ${\mathbf W}_{k}$, ${\mathbf W}_{v}$, ${\mathbf W}_f$, and ${\mathbf b}_f$ are learnable parameters, and ``;" indicates the concatenation operation. The matrices ${\mathbf Q}_{F_c}$, ${\mathbf K}_{F_c}$, ${\mathbf V}_{F_c}$ are known as the \textit{query}, \textit{key}, and \textit{value} in self-attention calculation, and here they are the linear transformation of the concatenated feature ${\mathbf F}_{c}$.

\subsection{Answer Querying Decoder}
Given an input image question pair, among a set of answers of interest, our Answer Querying Decoder predicts whether each candidate answer matches the corresponding image question pair and uses the candidate with the highest probability as the final answer. For this purpose, we employ a two-layer transformer decoder followed by a linear projector as our classifier, and introduce a set of learnable candidate answer embeddings together with the fused image-question feature ${\mathbf F}_f$ as the input of the decoder. 
%Answers could be either an open answer  (e.g. Pulmonary nodules, right temporal lobe, Xray, etc.) or a closed answer (Yes, No, etc). 
Assuming there are $C$ answer classes in total, we need $C$ candidate answer embeddings with one-to-one correspondence to the $C$ answer classes. These answer embeddings, collectively represented by a matrix ${\mathbf A}$, are randomly initialised and will be updated during training through a self-attention module, a cross-attention module, and a feed-forward network(FFN) in order. Both the self-attention module and the cross-attention module implement the multi-head self-attention ($MSA$(\textit{query}, \textit{key}, \textit{value})) but with different \textit{key}, \textit{query}, and \textit{value}. The self-attention module computes the relation between different answer embeddings by using ${\mathbf A}$ to construct all the \textit{key}, \textit{query}, and \textit{value} matrices. The cross-attention module cares about the relation between the answer embeddings ${\mathbf A}$ and the fused image-question feature ${\mathbf F}_f$. It thus uses the answer embedding ${\mathbf A}$ as the \textit{query} and the fused image-question feature ${\mathbf F}_f$ as the \textit{key} and \textit{value} to compute the attention and further updates the answer embeddings by combining the attended image-question features. Mathematically, denoting the answer embeddings at the $l$-th layer as ${\mathbf A}_l$, it will be updated from the output of the previous layer ${\mathbf A}_{l-1}$ as follows:
\begin{align}
&{\mathbf A}_l=MSA({\mathbf A}_{l-1}, {\mathbf A}_{l-1}, {\mathbf A}_{l-1})\\\nonumber
& {\mathbf A}_l=MSA({\mathbf A}_{l}, {\mathbf F}_{f}, {\mathbf F}_{f})\\\nonumber
&{\mathbf A}_l=FFN({\mathbf A}_l),
\end{align}
where $l=1 \cdots L$ and $L$ is the number of Transformer decoder layers. Through this process, the image-question features are injected into the answer embeddings and used to refine the latter. The refined $C$ answer embeddings are sent to the final linear projection layer followed by a sigmoid function $\sigma(\cdot)$ to predict the probabilities of answer classes. That is:
\begin{equation}
\mathbf {p} = \sigma({\mathbf W}_{A} {\mathbf A_L}  + {\mathbf b}),
\end{equation}
where ${\mathbf W}_{A}$ and ${\mathbf b}$ are learnable parameters, and ${\mathbf p}$ is a vector comprising of $C$ probabilities corresponding to $C$ answer classes. The answer class with the highest probability is chosen as the predicted answer.
%we denle use the multiote ion moduthem as y = [y1, ..., yN ], where $yN \in \left \{ 0, 1 \right \} $, N = 1, ..., N, is a discrete binary indicator. yN = 1 if the corresponding image question pair has the N-th answer, otherwise yN = 0.  Using image question fusion features as input, the algorithm uses every answer embedding to query fusion features. The queried feature vectors $\mathbf Q \in \mathbb{R}^{N \times D}$for N candidate answers at the last layer. Our model predicts the probabilities of the presence of each answer. 

%We use the standard Transformer architecture, which has a self-attention module, a cross-attention module, and a position-wise feed-forward network(FFN). Each Transformer decoder layer i updates the queries $Qi - 1$  from the output of its previous layer as follows: 

%\begin{equation}
 %   \mathbf {Attention(Q,K,V)} = \mathbf { softmax(\frac{QK^T}{\sqrt{d_k}})V} \\
%\end{equation}
%we get the queried feature vectors Q for N answers at the last layer. To perform multi-label classification, we treat each answer prediction as a binary classification task and project the feature of each class Q to a logit value using a linear projection layer:
%\begin{equation}
%\mathbf {p\left ( n \right )} = \mathbf {W\left ( n \right ) Q\left ( n \right )  + b\left ( n \right ) }
%\end{equation}
%Where W(n) and b(n) is the learnable weight and bias, and the Q(n) is the attention map.
% \begin{figure}
% \includegraphics[width=\textwidth]{fig2.png}
% \caption{The Structure of Our Decoder} \label{fig2}
% \end{figure}

\subsection{Loss Function} 
%Because we used the built-in cross-attention mechanism in transformer decoders, the binary cross entropy loss and focal loss both behave well in our framework. 
Medical VQA faces a significant class imbalance problem:  Yes/No answer classes are much larger than long open answer classes.  
In order to address the sample imbalance problem more effectively, we choose a simplified asymmetric loss, which is a variant of focal loss while the hyper-parameter $\mathbf{\gamma}  $ is set differently for positive and negative classes, as shown in Eqn.~\ref{Eqn:loss}:
%Our framework can predict the answer probabilities $ \mathbf{p} \in \left \{ \mathbf{N} \times \mathbf{D'} \right \} $ of a fusion feature. Then we use the following loss function to calculate the loss:

\begin{equation}\label{Eqn:loss}
\mathcal{L}=\frac{1}{C} \sum_{c=1}^{C}\left\{\begin{array}{ll}
\left(1-p_{c}\right)^{\gamma+} \log \left(p_{c}\right), & y_{c}=1 \\
\left(p_{c}\right)^{\gamma-} \log \left(1-p_{c}\right), & y_{c}=0
\end{array}\right.
\end{equation}
where $y_c$ is the ground-truth binary label, indicating if the input image-question pair has the answer class $c$, while $p_c$ is the predicted probability for the class $c$. The total loss is computed by averaging this loss over all samples in the training data set. We set the hyper-parameters $\gamma + = 1$ and$\gamma - = 4$ by default.

%----------------------------------------------------------\-------------------------------------------------------
\section{Experiments And Results}
\subsection{Datasets}
We conduct our experiments on two medical VQA benchmarks: VQA-RAD \cite{lau2018dataset} and PathVQA \cite{he2020pathvqa}, which are described as follows.

\noindent\textbf{VQA-RAD} is the most commonly used radiology dataset seen to date, containing 315 images and 3515 question-answer pairs, each corresponding to at least one question-answer pair. The types of questions include 11 categories: ``anomalies", ``properties", ``color", ``number", ``morphology", ``organ type", ``other", and ``section". 58$\%$ of the questions are close-end questions and the rest are open-end questions. The images are of the body's head, chest, and abdomen. Manual division of the training and test sets is required. For comparability, we divide the data set according to the MMQ method\cite{do2021multiple}.

\noindent\textbf{PathVQA} is a dataset for exploring pathology VQA. Images with captions were extracted from digital resources (electronic textbooks and online libraries). Open-end questions account for 50.2$\%$ of all questions. For the closed-end yes/no questions, the answers are balanced with 8,145 yes and 8,189 no questions. PathVQA consists of 32,799 question-answer pairs, 1,670 pathology images collected from two pathology textbooks, and 3,328 pathology images collected from the PEIR digital library \cite{PEIR}. For comparability, we also divide the data set according to the MMQ method \cite{do2021multiple}.

% \subsection{Implementation Details}
% For VQA-RAD, We train our model for 50 Epoch with 2 decoder layers, and our answer embedding dimension set to 2048 for VQA-RAD. In table 1, we get an accuracy of 79.19$\%$(19.09$\%$ absolute improvement) on Open-ended questions, 81.2$\%$(1.2$\%$ absolute improvement) closed questions, and 80.48$\%$(8.48$\%$ absolute improvement) over-all questions. 
% For pathVQA dataset, We train our model for 50 Epoch with 2 decoder layers, and our answer embedding dimension set to 1024, because of GPU limitation. In table 1, We get an accuracy of 54.85$\%$(41.45$\%$ absolute improvement) on Free-form questions, 88.85$\%$(4.85$\%$ absolute improvement) Yes/No questions, and 74.61(25.81$\%$ absolute improvement) over-all questions. The result will be better if increasing the answer embedding. We proved this in our ablation experiments.

%\subsection{Experiment results}
\subsection{Comparison with the state-of-the-art methods}
We compare our proposed model with 7 state-of-the-art (SOTA) Medical VQA approaches, including StAn~\cite{he2020pathvqa}, BiAn~\cite{he2020pathvqa}, MAML~\cite{finn2017model}, MEVF~\cite{nguyen2019overcoming}, MMQ~\cite{do2021multiple}, PubMedCLIP~\cite{eslami2021does}, and MMBERT~\cite{khare2021mmbert}. The first 6 methods are classification-based approaches. They are chosen because they are among the best performers on the two benchmarks VQA-RAD and PathVQA. The last method MMBERT~\cite{khare2021mmbert} is chosen as a representative of generation-based approaches, which has the reported performance on VQA-RAD. Except PubMedCLIP~\cite{eslami2021does} and MMBERT~\cite{khare2021mmbert} whose results are quoted from their original papers, the results of other methods are quoted from MMQ~\cite{do2021multiple}. It is noted that same as our Q2ATransformer, PubMedCLIP~\cite{eslami2021does} and MMBERT~\cite{khare2021mmbert} employ the same data split as MMQ~\cite{do2021multiple}. Therefore these results are strictly comparable.

As shown in Table~\ref{tab1}, on both datasets, our Q2ATransformer consistently outperforms the compared models. Specifically, compared with the second best performer, on VQA-RAD, we achieve an accuracy of 79.19$\%$ (16.09$\%$ absolute improvement) on Open-end questions, 81.2$\%$ (1.2$\%$ absolute improvement) on close-end questions, and 80.48$\%$ (8.48$\%$ absolute improvemen) across all questions; on PathVQA, we achieve an accuracy of 54.85$\%$ (41.45$\%$ absolute improvement) on open-end questions, 88.85$\%$ (4.85$\%$ absolute improvement) on Yes/No questions, and 74.61$\%$ (25.81$\%$ absolute improvement) across all questions. The results could be even better if we increase the dimension of the candidate answer embeddings, as shown in our ablation experiments. From these results, we can see our Q2ATransformer demonstrates overwhelming advantages on open-end questions, which supports our analysis that by interacting answer information with fused image-question features, our model could better tackle long answer questions. Our model also outperforms the generation-based method MMBERT~\cite{khare2021mmbert}, since we reduce the search space of answers while MMBERT~\cite{khare2021mmbert} could generate non-existent answers.

\begin{table}
\centering
\caption{Performence comparison of different methods. $\dagger$ and $\ddagger$ indicate the methods are classification-based(closed-type) or generation-based(open-type), respectively.}
\label{tab1}
\begin{tabular}{c|c|c|c|c|c|c|c} 
\hline
\multirow{2}{*}{\begin{tabular}[c]{@{}c@{}}References \\ Methods\end{tabular}} & \multirow{2}{*}{\begin{tabular}[c]{@{}c@{}}Fusion \\ Methods\end{tabular}} & \multicolumn{3}{c|}{PathVQA}                                                                                                                                   & \multicolumn{3}{c}{VQA-RAD}                                                                                                                                           \\ 
\cline{3-8}
                                                                               &                                                                            & \begin{tabular}[c]{@{}c@{}}Free-form\end{tabular} & \begin{tabular}[c]{@{}c@{}}Yes/No\end{tabular} & \begin{tabular}[c]{@{}c@{}}Over-all\end{tabular} & \begin{tabular}[c]{@{}c@{}}Open-ended\end{tabular} & \begin{tabular}[c]{@{}c@{}}Close-ended\end{tabular} & \begin{tabular}[c]{@{}c@{}}Over-all\end{tabular}  \\ 
\hline
StAn$^\dagger$\cite{he2020pathvqa}                                                                           & SAN                                                                        & 1.6                                                  & 59.4                                              & 30.5                                                & 24.2                                                  & 57.2                                                   & 44.2                                                 \\ 
\hline
BiAn$^\dagger$\cite{he2020pathvqa}                                                                           & BAN                                                                        & 2.9                                                  & 68.2                                              & 35.6                                                & 28.4                                                  & 67.9                                                   & 52.3                                                 \\ 
\hline
\multirow{2}{*}{MAML$^\dagger$\cite{finn2017model}}                                                          & SAN                                                                        & 5.4                                                  & 75.3                                              & 40.5                                                & 38.2                                                  & 69.7                                                   & 57.1                                                 \\ 
\cline{2-8}
                                                                               & BAN                                                                        & 5.9                                                  & 79.5                                              & 42.9                                                & 40.1                                                  & 72.4                                                   & 59.6                                                 \\ 
\hline
\multirow{2}{*}{MEVF$^\dagger$\cite{nguyen2019overcoming}}                                                          & SAN                                                                        & 6.0                                                  & 81.0                                              & 43.6                                                & 40.7                                                  & 74.1                                                   & 60.7                                                 \\ 
\cline{2-8}
                                                                               & BAN                                                                        & 8.1                                                  & 81.4                                              & 44.8                                                & 43.9                                                  & 75.1                                                   & 62.7                                                 \\ 
\hline
\multirow{2}{*}{MMQ$^\dagger$\cite{do2021multiple}}                                                           & SAN                                                                        & 11.2                                                 & 82.7                                              & 47.1                                                & 46.3                                                  & 75.7                                                   & 64.0                                                 \\ 
\cline{2-8}
                                                                               & BAN                                                                        & 13.4                                                 & 84.0                                              & 48.8                                                & 53.7                                                  & 75.8                                                   & 67.0                                                 \\ 
\hline
PubMedCLIP$^\dagger$\cite{eslami2021does}                                                                     & \--                                                                          & \--                                                    & \--                                                 & \--                                                   & 60.1                                                  & 80                                                     & 72.1                                                 \\ 
\hline
MMBERT$^\dagger$\cite{khare2021mmbert}                                                                           & \--                                                                       & \--                                                & \--                                             & \--                                               & 63.1                                                 & 77.9                                                   & 72.0                                                \\
\hline
Ours                                                                           &                                                                        & \textbf{54.85}                                                & \textbf{88.85}                                             & \textbf{74.61}                                               & \textbf{79.19}                                                 & \textbf{81.2}                                                   & \textbf{80.48}                                                \\
\hline
\end{tabular}
\end{table}

\begin{table}~\label{tab:abalation_study}
\centering
\caption{Ablation Studies. BAN, SAN, and CMAN stand for Bilinear Attention Network~\cite{kim2018bilinear}, Stacked Attention Network~\cite{yang2016stacked} and ours Cross-modality Attention Network, respectively; Decoder refers to our Answer-Querying Decoder. }
\label{tab:ablation_study}
\setlength{\tabcolsep}{1.2mm}
\begin{tabular}{c|cccc|ccc|ccc} 
\hline
\multirow{2}{*}{\#} & \multirow{2}{*}{BAN}                                & \multirow{2}{*}{SAN} & \multirow{2}{*}{CMAN} & \multirow{2}{*}{Decoder} & \multicolumn{3}{c|}{VQA-RAD} & \multicolumn{3}{c}{PathVQA}                                                        \\ 
\cline{6-11}
                    &                                                     &                      &                      &                          & open  & closed & overall     & free-form                      & yes/no                    & overall                    \\ 
\hline
1                   & \begin{tabular}[c]{@{}c@{}}\Checkmark\\\end{tabular} &            &            &                & 43.62  & 75.56  & 64.1            & \multicolumn{1}{c}{15.03}      & \multicolumn{1}{c}{78.24}      & \multicolumn{1}{c}{51.69}       \\
2                   & \Checkmark                                           &            &            & \Checkmark                & 54.36 & 80.07  & 70.84       & 44.78                     & 88.29                  & 70.09                      \\
3                   &                                           & \Checkmark            &            &                & 61.07 & 77.07  & 71.33       & \multicolumn{1}{c}{44.58}      & \multicolumn{1}{c}{86.29}      & \multicolumn{1}{c}{68.88}       \\
4                   &                                           & \Checkmark            &            & \Checkmark                & 73.83 & 80.08  & 77.83       & \multicolumn{1}{c}{52.88} & \multicolumn{1}{c}{88.44} & \multicolumn{1}{c}{73.51}  \\
5                   &                                           &            & \Checkmark            &                & 69.13 & 76.32  & 73.73       & 47.53                     & 86.73                     & 70.31                      \\
6                   &                                           &            & \Checkmark            & \Checkmark                & 79.19 & 81.2   & 80.48       & 54.85                     & 88.85                     & 74.61                      \\
\hline
\end{tabular}
\end{table}

\vspace{-6mm}
\subsection{Ablation study}

To investigate the contributions of our proposed feature fusion module CMAN and the decoder for answer querying, we conduct extensive ablation studies to compare different configurations of our model, as presented in Table.~\ref{tab:abalation_study}. Here BAN, SAN, and CMAN are three attention networks to fuse image and question features, representing Bilinear Attention Network~\cite{kim2018bilinear}, Stacked Attention Network~\cite{yang2016stacked} and ours Cross-modality Attention Network, respectively; Decoder represents our Answer-Querying Decoder. The symbol \Checkmark indicates the inclusion of the corresponding component. All the experiments in Table.~\ref{tab:abalation_study} are performed based on the same image and question encoders.
%to clearly demonstrate the effect of our CTAN and Decoder.

% we experiment with the following baselines on the two benchmark datasets. The results are reported in Table.~\ref{tab:abalation_study}.

\noindent\textbf{Impact of the CMAN.} The benefit of using CMAN can be well reflected by the improvement from \#1 to \#5 or from \#3 to \#5 in Table.~\ref{tab:abalation_study}, indicating the effectiveness of our proposed CMAN over BAN and SAN for image and question feature fusion. This is because compared with BAN which multiplies the image and question features or SAN which does a direct matrix summation for fusion, our CMAN directly concatenates the two channels of features together and then calculates attention for fusion. Through this, our CMAN mitigates the information loss due to the multiplication or summation operation during feature fusion in BAN or SAN.
%The reason why BAN is not as effective as our fusion model is probably because BAN fuses the features of two channels together by multiplication of the two channels' features. However, we directly concatenate them together and then do attention, so that no information is lost. In contrast, BAN will lose some information.
%The dimension of our fusion model after fusion is the image width + text width. The reason why the effect of SAN is not as good as our fusion model may also be due to the fact that SAN does a direct matrix summation when fusing the features of two channels together, which will lose some information than directly concatenate them together and then do attention. 

% We replaced the fusion algorithm with BAN and SAN, The experimental results in table 4 prove that the effect is obviously not as good as using CTAN as fusion method.

\noindent\textbf{Contribution of the decoder.} As shown, the inclusion of our answer querying decoder could boost the model performance. To verify the robustness of our decoder, we incorporate it with three different attention modules shown in \#2, \#4 and \#6. By comparing \#2 to \#1, \#4 to \#3, or \#6 to \#5, it can be observed that our answer querying decoder can bring significant performance gain with all three attention mechanisms. Especially, when combining our CMAN and decoder, we can achieve the new SOTA results.
% If we remove the transformer decoder, and classify the result after fusion mode directly. The result in table 3 shows that the performance significantly dropped in the table. 

\noindent\textbf{Impact of answer embedding size.} The experimental results in Fig.3 show that as the dimension of answer embedding increases, the model's performance improves while the best result is obtained when the embedding size is around 2048. However, increasing the embedding size will also increase the computational cost, while the performance improvement becomes saturated. As a tradeoff, our model adopts 1024-dimensional answer embeddings.

\begin{figure}[t]
\centering
\includegraphics[width=0.9\textwidth]{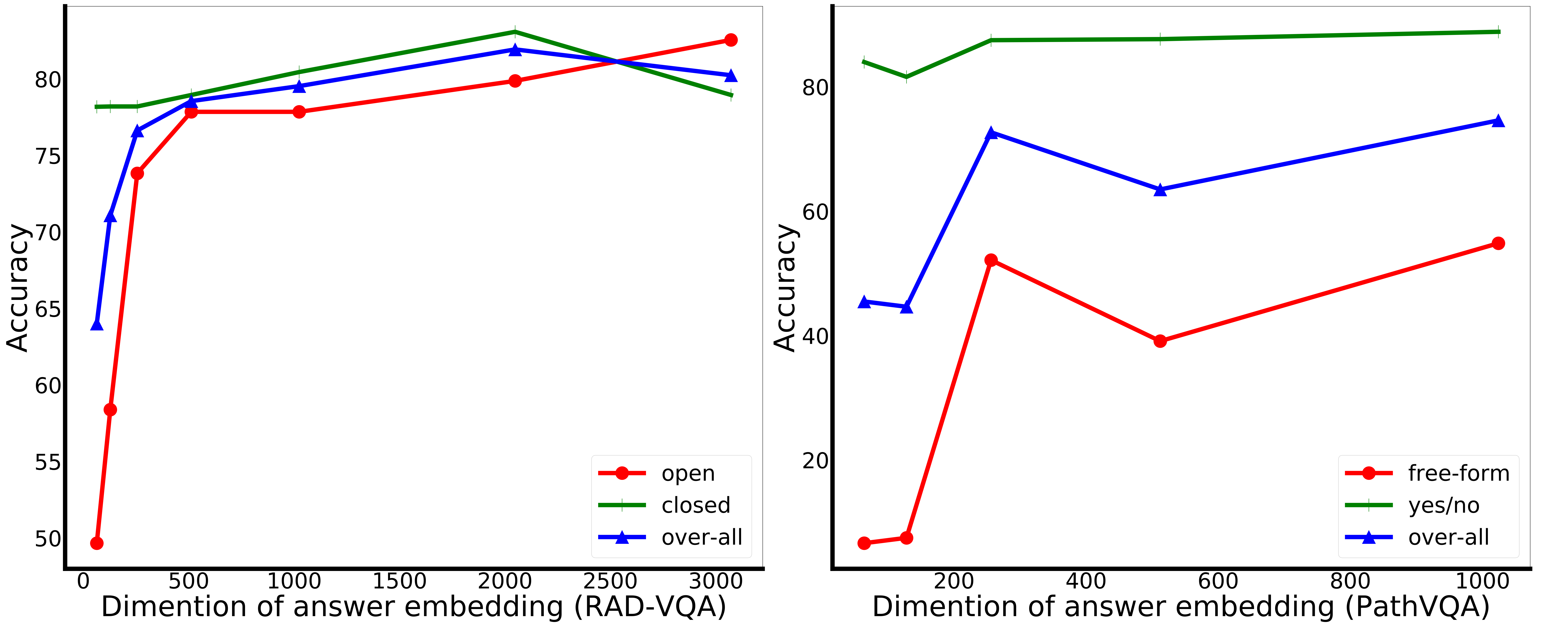}
\caption{Ablation study about different dimensions of answer embeddings.} \label{Dimention_change}
\end{figure}

\subsection{Qualitative results}
\begin{figure}[hp]
\centering
\includegraphics[width=0.95\textwidth]{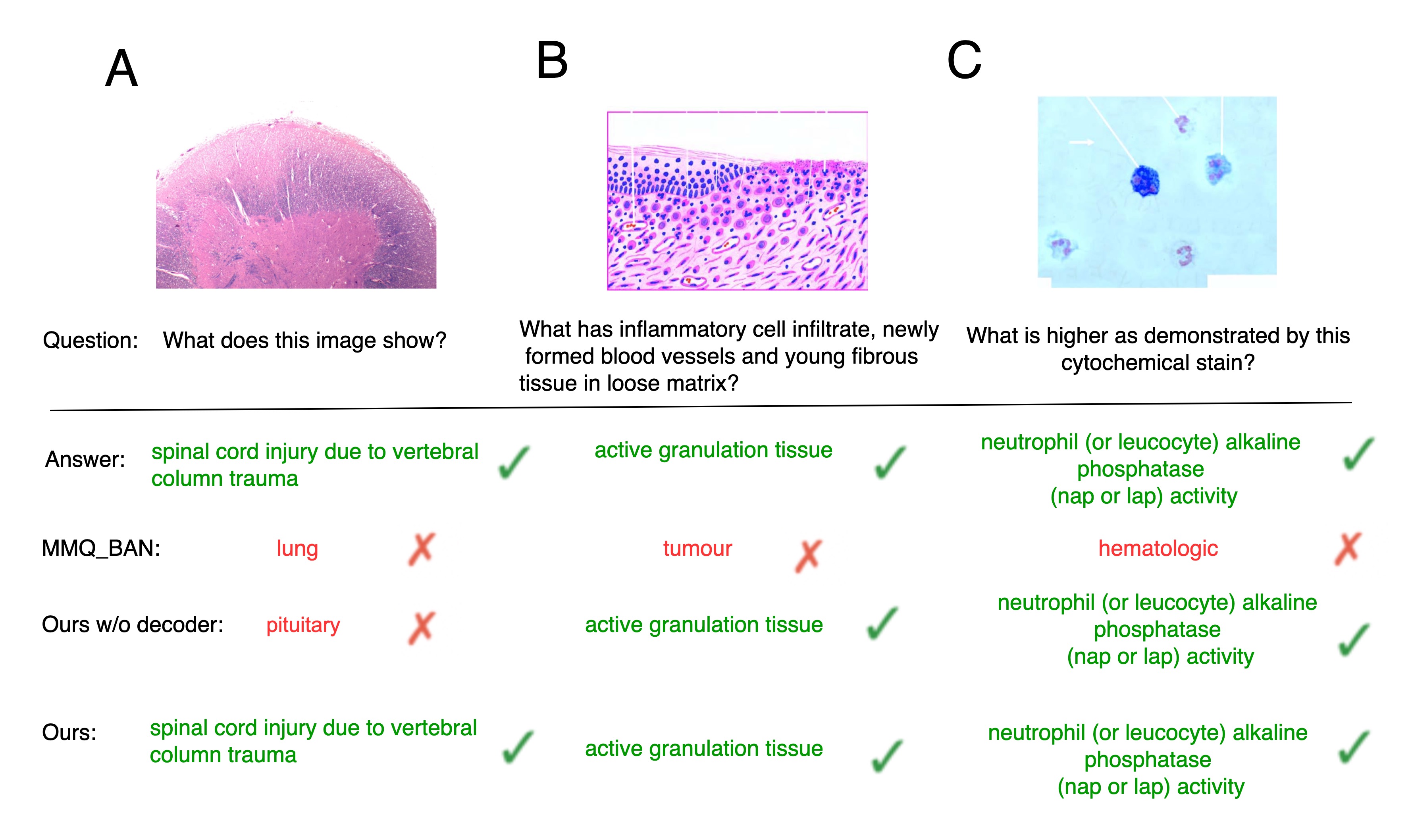}\\
\vspace{-6mm}
\caption{Example results from PathVQA dataset.} \label{examples-path}
\end{figure}

\begin{figure}[!hp]
\centering
\includegraphics[width=0.95\textwidth]{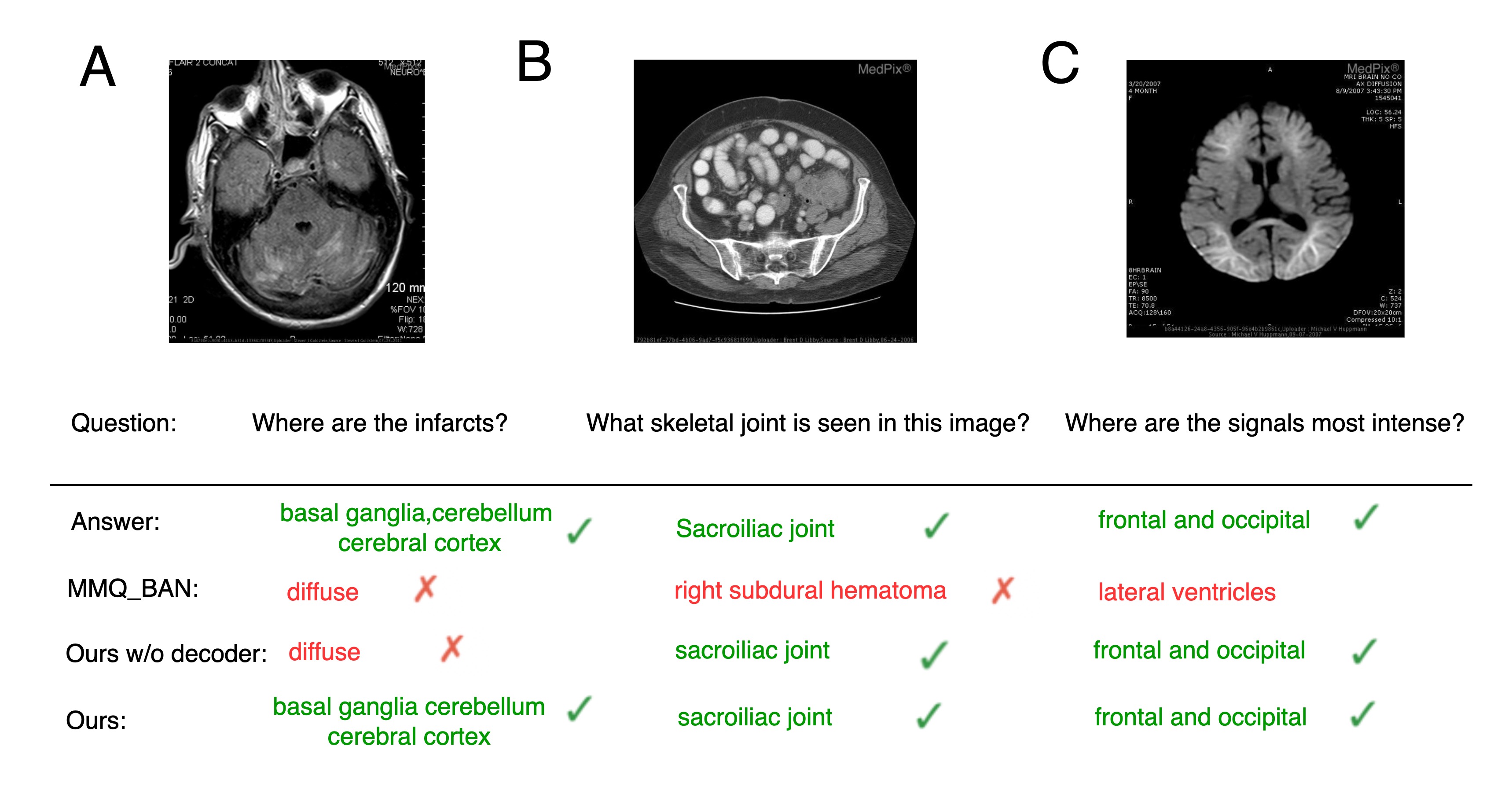}
\vspace{-6mm}
\caption{Example results from VQA-RAD dataset.} \label{examples-RAD}
\end{figure}

Example results from PathVQA and VQA-RAD datasets are given in Figure~\ref{examples-path} and Figure~\ref{examples-RAD}, respectively. As can be seen, for these examples where MMQ using BAN for feature fusion fails, our Q2ATransformer w/o decoder has been able to correct most of them using the proposed CMAN fusion module. The performance could be further improved with our Answer Querying Decoder by learning candidate answer embedding through their interactions with the fused image-question features.
% , and figure~\ref{examples_path} gives three example results from VQA-RAD, where 
%Figure 4 visualizes the results of our model with also another models on the PathVQA dataset, and our model is better at predicting accurate answers to longer questions than the others.
%
%

\subsection{Limitation and Discussion}
As described in Section~\ref{method}, we treat each answer as a learnable embedding and use all embeddings as the query to compute the attention map in our decoder. Since we compute the global self-attention, this may increase computation overhead when the number of answer classes is very large. This problem has been encountered in NLP  when processing long sequences. Some solutions have been proposed, such as dynamically computing sparse attention%~\cite{correia2019adaptively} 
, which can significantly reduce computational overhead and will be explored in our future work.

\section{Conclusion}
In this paper, we propose a semi-open framework for medical VQA, which successfully enrolls answer semantic information into the answer class prediction process through our designed mechanism to correlate the answering embeddings with the fused image-question features, which improves the accuracy significantly. It enriches the existing closed-type and open-type medical VQA frameworks and refreshes the SOTA performance on the two benchmarks, especially for the open-end questions. 

%It is noted that our framework is still classification-based approach and thus inherits the \textbf{limitation} of inflexibility to generate unseen questions, which exists with the current mainstream of medical VQA methods and should be investigated for relaxation in our future work.

% \begin{table}[]
% \caption{Result with no decoder.}\label{tab3}
% \setlength{\tabcolsep}{6mm}
% \begin{tabular}{c|c|c|c}
% \hline
%                         & Open      & Closed & Over-all \\ \hline
% VQA RAD Without Decoder & 69.13     & 76.32  & 73.73    \\ \hline
%                         & Free-form & Yes/No & Over-all \\ \hline
% PathVQA Without Decoder & 47.53     & 86.73  & 70.31    \\ \hline
% \end{tabular}
% \end{table}

% \begin{table}[h]
% \caption{Result with different fusion model.}\label{tab4}
% \setlength{\tabcolsep}{6mm}
% \begin{tabular}{c|c|c|c}
% \hline
%                  & Open           & Closed         & Over-all       \\ \hline
% VQA RAD With BAN & 54.36          & 80.07          & 70.84          \\ \hline
% VQA RAD With SAN & 73.83          & 80.08          & 77.83          \\ \hline
% Ours             & \textbf{79.19} & \textbf{81.2}  & \textbf{80.48} \\ \hline
%                  & Free-from      & Yes/No         & OVer-all       \\ \hline
% PathVQA with BAN & 44.78          & 88.29          & 70.09          \\ \hline
% PathVQA with SAN & 52.88          & 88.44          & 73.51          \\ \hline
% Ours             & \textbf{54.85} & \textbf{88.85} & \textbf{74.61} \\ \hline
% \end{tabular}
% \end{table}

%\newpage
\bibliographystyle{splncs04}
\bibliography{ref}
\end{document}